\documentclass{article}
\usepackage{styles/iclr2027/iclr2027_conference,times}
\usepackage[T1]{fontenc}
\usepackage{amsmath,amssymb,mathtools}
\usepackage{booktabs}
\usepackage{tabularx}
\usepackage{multirow}
\usepackage{float}
\usepackage{placeins}
\usepackage{needspace}
\usepackage{capt-of}
\usepackage{xspace}
\usepackage{microtype}
\usepackage{graphicx}
\usepackage{xcolor}
\usepackage{colortbl}
\usepackage{pifont}
\usepackage{url}
\usepackage[colorlinks=true,allcolors=blue]{hyperref}

\newcommand{\method}{\textsc{Vetta}\xspace}
\newcommand{\Atok}{A^{\mathrm{tok}}}
\newcommand{\Aturn}{A^{\mathrm{turn}}}
\newcommand{\Vtok}{V^{\mathrm{tok}}}
\newcommand{\Vturn}{V^{\mathrm{turn}}}

\newcolumntype{Y}{>{\raggedright\arraybackslash}X}
\newcolumntype{R}{>{\raggedleft\arraybackslash}X}

\newsavebox{\paperembeddedbox}
\newenvironment{embeddedpair}{%
  \par\begin{lrbox}{\paperembeddedbox}\begin{minipage}{\linewidth}%
}{%
  \end{minipage}\end{lrbox}%
  \Needspace{\dimexpr\ht\paperembeddedbox+\dp\paperembeddedbox\relax}%
  \noindent\usebox{\paperembeddedbox}\par
}

\title{VETTA: Coordinating Turn- and\\
Token-Level Credit Assignment for\\
Multi-Turn LLM Agents}
\author{
Jiaju Chen\textsuperscript{1,2} \quad
Min Yang\textsuperscript{3,2} \quad
Jinghua Piao\textsuperscript{4,2}\thanks{Corresponding authors} \\
\bfseries Xiaochong Lan\textsuperscript{4} \quad
Xu Xia\textsuperscript{5,2} \quad
Xiangnan He\textsuperscript{1} \quad
Yong Li\textsuperscript{4,2}\footnotemark[1] \\[0.6em]
\textsuperscript{1}University of Science and Technology of China \quad
\textsuperscript{2}Zhongguancun Academy \\
\textsuperscript{3}Shandong University \quad
\textsuperscript{4}Tsinghua University \quad
\textsuperscript{5}Southeast University \\[0.6em]
\texttt{cjj01@mail.ustc.edu.cn} \quad
\texttt{minyang@mail.sdu.edu.cn} \\
\texttt{Pjh22@mail.tsinghua.edu.cn} \quad
\texttt{lanxc22@mail.tsinghua.edu.cn} \\
\texttt{s-xx25@bza.edu.cn} \quad
\texttt{xiangnanhe@gmail.com} \\
\texttt{liyong07@tsinghua.edu.cn}}

\iclrfinalcopy

\newcommand{\codeavailability}{Code is available at
\url{https://github.com/Jiaju-Chen/VETTA-official}.}

\begin{document}
\maketitle
\lhead{}

\begin{abstract}
Multi-turn LLM agents often receive sparse task feedback across several interactions, while generating each response token by token. 
This creates two related credit-assignment questions: which responses helped achieve the outcome, and which generation decisions mattered within each response? Existing methods typically focus on only one level: turn-level methods evaluate complete responses but do not distinguish the decisions within them; token-level methods can propagate feedback across turns but do not explicitly model credit for each response.
These complementary limitations motivate learning credit at both levels and coordinating it in a single policy update.
We introduce \method, a credit assignment method that jointly learns turn- and token-level
values through separate heads on a shared lightweight critic. \method computes
advantages along both temporal sequences and combines each turn advantage
with a within-response-centered token residual for PPO updates. Furthermore, to reduce
value-learning cost, the critic retains only early Transformer blocks from
the pretrained checkpoint used to initialize the actor. 
Experiments on four agent benchmarks demonstrate VETTA's strong performance.
With Qwen2.5-7B-Instruct, VETTA achieves success rates of 95.5\% on ALFWorld,
76.0\% on WebShop, and 52.2\% on ScienceWorld. With
Qwen2.5-VL-3B-Instruct, it reaches 84.9\% and 80.7\% success on Sokoban
under sparse and dense rewards, respectively.
Critic-depth comparisons further show strong task
performance with substantially lower critic-side computation.
These results demonstrate that coordinating turn- and token-level credit
with a lightweight shared critic improves agent performance while keeping
value estimation efficient.
\codeavailability
\end{abstract}

\section{Introduction}
\label{sec:introduction}

Large language models (LLMs) have demonstrated capabilities in reasoning,
tool use, and interactive decision-making, enabling agents to pursue goals
through interaction with external environments
\citep{yao2022react,schick2023toolformer}.
Reinforcement learning (RL) has further improved these capabilities by
optimizing LLM policies from task feedback
\citep{shao2024deepseekmath,deepseekai2025deepseekr1,jin2025searchr1,feng2025gigpo}.
Many agentic tasks, including operating applications, navigating shopping
websites, and carrying out household instructions
\citep{trivedi2024appworld,yao2022webshop,shridhar2021alfworld}, require
repeated action--observation rounds, while rewards are
often sparse or delayed \citep{zhou2024archer,peng2026hiper}.
Learning from such feedback requires distinguishing the contributions of
individual decisions across an extended interaction
\citep{feng2025gigpo,cheng2026graphgpo}.

However, commonly used RL approaches face challenges at different
granularities: a direct multi-turn application of outcome-based
GRPO~\citep{shao2024deepseekmath,feng2025gigpo} assigns the same
trajectory-level relative advantage to every turn, whereas token-level
PPO~\citep{schulman2017ppo,li2025turnppo,hou2026sao} must estimate values
for generated-token decisions from delayed feedback across successive
interactions.

Recent work has therefore introduced turn-level credit into both optimization
paradigms, treating each complete response action as a unit for advantage
estimation. Critic-free methods estimate action-level advantages from
sampled-rollout returns
\citep{feng2025gigpo,zhu2026gagpo,he2026hgpo}, or from constructed
state-transition graphs
\citep{wang2026group,cheng2026graphgpo}. Critic-based methods such as Turn-PPO~\citep{li2025turnppo}
instead perform value and advantage estimation once per turn, treating the
complete response as a whole. These approaches distinguish credit across actions 
but leave generation decisions within each action undifferentiated, 
suggesting that turn- and token-level credit could be coordinated to 
guide learning at both scales.

\begin{figure}[!t]
  \centering
  \includegraphics[width=\linewidth]{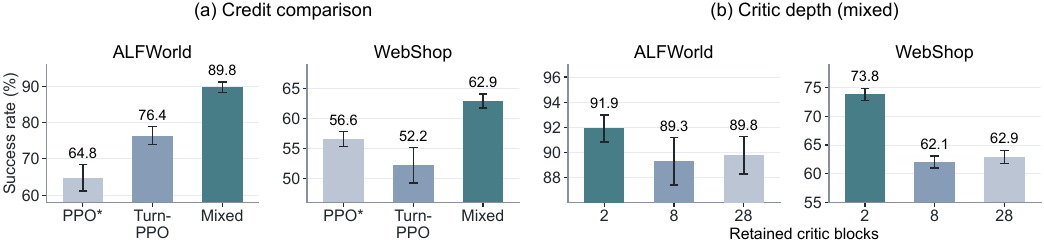}
  \caption{Success rates on ALFWorld and WebShop with Qwen2.5-1.5B-Instruct:
  (a) full-depth PPO*, Turn-PPO, and mixed
  configurations; (b) the mixed configuration with 2, 8, or 28 critic blocks.
  Bars show means and error bars show sample standard deviations over three
  decoding seeds for each checkpoint.}
  \label{fig:intro-observations}
\end{figure}

To examine whether combining turn- and token-level credit improves policy
learning, we compare PPO* (token-level credit across turns), Turn-PPO, and a
mixed configuration that combines advantages from both granularities
(Section~\ref{sec:advantage}).
As shown in Figure~\ref{fig:intro-observations}(a), the mixed
configuration achieves higher mean success rates than either alternative on both
ALFWorld and WebShop. These results suggest that turn- and token-level credit
can provide complementary information for multi-turn policy learning.

Beyond the choice of credit signal, critic-based PPO adds value inference and
critic updates to the training workload. We therefore examine whether
simplifying the critic can reduce this cost without sacrificing policy
performance. With mixed credit, Figure~\ref{fig:intro-observations}(b)
compares critics retaining 2, 8, or all 28 Transformer blocks. On both
ALFWorld and WebShop, the two-block critic achieves a higher mean success
rate than the full-depth critic. These results suggest that the critic's full
pretrained depth is not necessary for strong policy performance in these
settings.

Building on these observations, we propose \method{} (\textbf{V}alue
\textbf{E}stimation at \textbf{T}urn and \textbf{T}oken levels for LLM
\textbf{A}gents), a framework for credit assignment in multi-turn interactions.
To model the two decision points, \method uses separate turn- and token-value
heads: the former estimates the value before a response is generated, while
the latter estimates values before its generated tokens. Both heads share a
critic backbone but learn from their respective return targets. We
compute generalized advantage estimates along the turn and generated-token
trajectories. Direct addition would let both advantages shift the overall
credit assigned to a response. Instead, we use the turn advantage as the
response-level signal and add a scaled, within-response-centered token
residual to distinguish generation decisions. The fused advantage guides
token-wise PPO updates. We further design a lightweight shared critic by
retaining only the early Transformer blocks from the pretrained checkpoint
used to initialize the actor, reducing the computation required to learn
both value estimates.

We evaluate \method on four multi-turn agent benchmarks:
ALFWorld~\citep{shridhar2021alfworld},
WebShop~\citep{yao2022webshop},
Sokoban~\citep{gymsokoban} and ScienceWorld~\citep{wang2022scienceworld}.
With Qwen2.5-7B-Instruct~\citep{qwen2025qwen25}, \method achieves success
rates of 95.5\% on ALFWorld, 76.0\% on WebShop, and 52.2\% on ScienceWorld.
With Qwen2.5-VL-3B-Instruct~\citep{qwen2025vl3b}, it reaches 84.9\% and
80.7\% success on Sokoban under sparse and dense rewards, respectively.
Across all four benchmarks, \method outperforms both critic-based and
critic-free baselines in overall success rate. In our depth
comparisons, retaining two rather than 28 critic blocks improves success
rates and reduces median critic-side time by approximately 87\% and 91\%
on ALFWorld and WebShop, respectively.

Our contributions are threefold:

\begingroup
\setlength{\leftmargini}{1em}
\begin{itemize}
    \item We formulate multi-turn credit assignment around two complementary
    roles: turn-level advantages provide credit for complete responses across
    interactions, while token-level advantages distinguish generation
    decisions within each response.
    \item We propose \method, which learns the two value estimates with
    separately supervised heads on a lightweight shared critic and combines
    their advantages through within-response centering before token-wise PPO
    updates.
    \item We evaluate \method on four agent benchmarks, demonstrating strong
    overall task success across different interaction environments and reward
    settings, and analyze the effects of advantage composition, critic sharing,
    and backbone depth.
\end{itemize}
\endgroup

\section{Preliminaries}
\label{sec:prelim}

In this section, we formalize the response- and token-level decision sequences used for value
estimation and policy optimization.

\subsection{Multi-turn LLM agent interaction}

Consider one finite interaction trajectory containing $K$ environment turns.
At turn $k\in\{1,\ldots,K\}$, the agent receives the context $H_k$, which
contains the task instruction, the current observation, and the preceding
interaction history included in the model context. Observations can be
textual or visual, depending on the benchmark. Conditioned on $H_k$,
the policy $\pi_\theta$, with parameters $\theta$, generates a
response $a_k=(y_{k,1},\ldots,y_{k,L_k})$, where $y_{k,j}$ is the $j$-th
generated token and $L_k$ is the number of valid generated tokens in that
response (the token mask is specified in Appendix~\ref{app:valid-tokens}).
After the response is completed, the environment executes the action
specified by $a_k$, returns a scalar reward $r_k$, and provides the next
observation $o_{k+1}$. The next context
$H_{k+1}$ incorporates that observation.  The same interaction can therefore
be viewed as a sequence of complete response decisions or as a finer sequence
of token-generation decisions.

\subsection{PPO and generalized advantage estimation}

For a generic decision sequence
$\tau=(s_t,a_t,r_t)_{t=0}^{T-1}$ containing $T$ decisions, let $t$ index
the policy decisions, where $s_t$, $a_t$, and $r_t$ denote the decision
context, action, and reward at step $t$.
With discount factor $\gamma\in[0,1]$, the standard policy objective is
\begin{equation}
  J(\theta)
  =\mathbb E_{\tau\sim\pi_\theta}
    \left[\sum_{t=0}^{T-1}\gamma^t r_t\right].
  \label{eq:discounted-return}
\end{equation}
A multi-turn interaction admits turn- and token-level sequences for value and
advantage estimation: a turn-level decision corresponds to a complete response
$a_k$, whereas a token-level decision corresponds to a generated token
$y_{k,j}$.

For a policy $\pi$, $Q^\pi(s_t,a_t)$ denotes the expected discounted return
after taking action $a_t$ in state $s_t$ and subsequently following $\pi$,
while $V^\pi(s_t)$ is the expected return before choosing that action. The
advantage, $A^\pi(s_t,a_t)=Q^\pi(s_t,a_t)-V^\pi(s_t)$, measures the action's
value relative to the policy's average at that state. Generalized advantage
estimation (GAE) uses rewards and value predictions along a sampled trajectory
to estimate this advantage~\citep{schulman2015gae}. PPO uses the estimated
advantage to guide policy updates, with probability-ratio clipping limiting
the incentive for large policy changes~\citep{schulman2017ppo}. VETTA retains
token-wise policy optimization and uses the two decision sequences to
construct complementary advantage estimates.

\section{Method}
\label{sec:method}

\begin{figure}[!t]
  \centering
  \includegraphics[width=\linewidth]{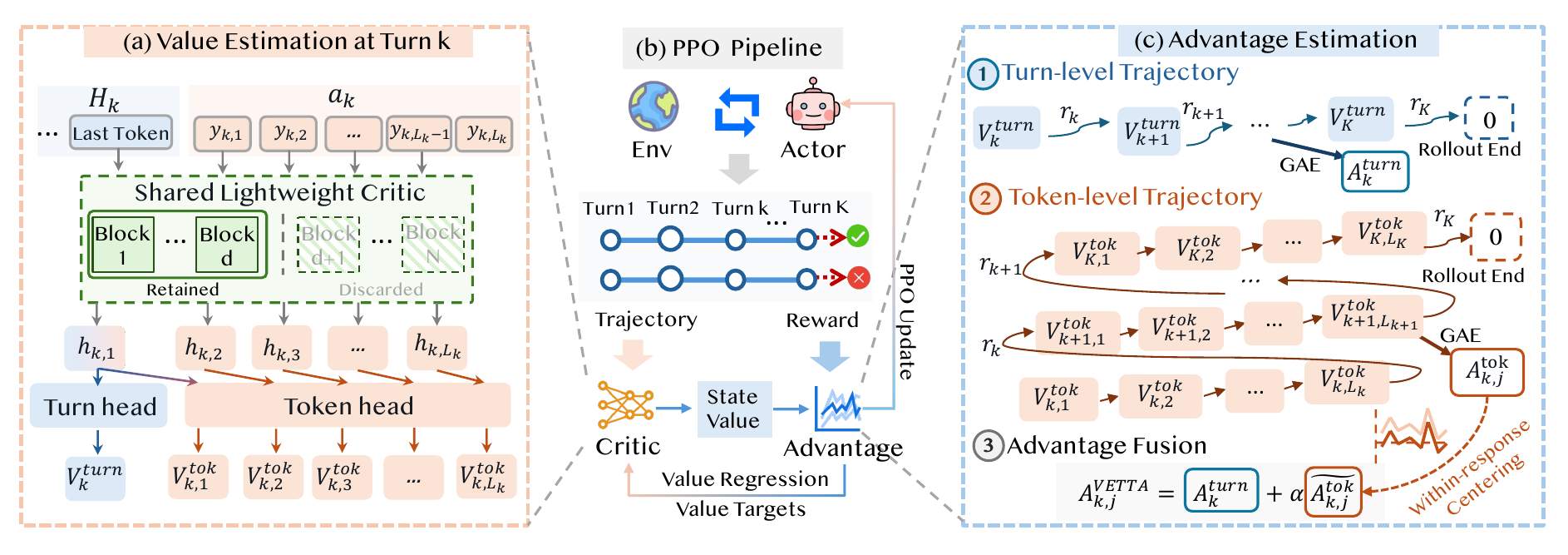}
  \caption{Overview of VETTA. (a) A shared critic retains early pretrained
  Transformer blocks and uses separate heads for turn- and token-level value
  estimation. (b) The actor interacts with the environment to collect
  trajectories and rewards; value estimates support advantage computation,
  followed by critic regression and PPO policy updates. (c) The collected
  trajectory is viewed at turn and token levels for GAE; token advantages are
  centered within each response and combined with its turn advantage. The implicit
  successor value and advantage are set to zero at rollout end.}
  \label{fig:method-overview}
\end{figure}

As a critic-based method, VETTA follows the PPO pipeline in
Figure~\ref{fig:method-overview}(b):
the actor collects multi-turn trajectories, and the critic provides value
estimates for advantage computation and policy optimization. Within this
pipeline, VETTA uses a shared lightweight critic to estimate values at both
turn and token levels (Figure~\ref{fig:method-overview}(a)), then combines the
resulting advantages to guide the actor update
(Figure~\ref{fig:method-overview}(c)). The two value heads are trained with
their respective return targets.

\subsection{Dual-granularity value estimation with a lightweight critic}
\label{sec:value-model}

At each environment turn, VETTA estimates future returns at two decision
granularities.  Before the response $a_k$ is generated, the turn-level value
$\Vturn_k$ evaluates the context $H_k$ and provides one value for the complete
response decision.  Within that response, the token-level value
$\Vtok_{k,j}$ evaluates the prefix available before generating $y_{k,j}$,
providing one value for each valid generated position.
Figure~\ref{fig:method-overview}(a) illustrates how the two sets of values are
read from a shared critic with causal attention.

In a single forward pass, the critic processes the context followed by the
sampled response $a_k=(y_{k,1},\ldots,y_{k,L_k})$ using a causal attention
mask. We write
$y_{k,<j}=(y_{k,1},\ldots,y_{k,j-1})$ for the response prefix available before
position $j$, with $y_{k,<1}=\emptyset$.  Let $h_{k,1}$ be the representation
at the final context position, and let $h_{k,j}$ for $j>1$ be the
representation at $y_{k,j-1}$.  Under causal masking, $h_{k,j}$ therefore
depends only on $(H_k,y_{k,<j})$ and is aligned with the decision to generate
$y_{k,j}$.

The two output heads map these representations to
\begin{equation}
  \begin{aligned}
    \Vturn_k=\Vturn(H_k)
      &=w_{\mathrm{turn}}^\top h_{k,1}+b_{\mathrm{turn}},\\
    \Vtok_{k,j}=\Vtok(H_k,y_{k,<j})
      &=w_{\mathrm{tok}}^\top h_{k,j}+b_{\mathrm{tok}},
  \end{aligned}
  \label{eq:value-model}
\end{equation}
where $w_{\mathrm{tok}}$ and $w_{\mathrm{turn}}$ are head-specific weight
vectors and $b_{\mathrm{tok}}$ and $b_{\mathrm{turn}}$ are their biases.
The two value heads are newly initialized and trained with separate
supervision targets. They have separate output parameters and use
representations from the same backbone. Both read $h_{k,1}$, while the token-value head additionally
reads the representations preceding subsequent generated tokens.

\noindent\textbf{Lightweight critic backbone.}
To limit critic cost, we construct its backbone from the pretrained checkpoint
used to initialize the actor. Of its $N$ Transformer blocks, we retain the
token embedding, the first $d\leq N$ blocks, and the final root mean square
normalization (RMSNorm), discarding the remaining blocks
(Figure~\ref{fig:method-overview}(a)). All retained critic components remain
trainable. This reduces the number of blocks used in critic forward and
backward passes without removing either value head. We examine the effect of
retained depth on policy performance and critic-side computation in
Section~\ref{sec:further-analysis}.

\subsection{Advantage estimation}
\label{sec:advantage}

VETTA applies standard GAE separately at the turn and token levels. The same
scalar reward $r_k$ is represented at both granularities: at the turn level,
it is assigned to turn $k$; at the token level, it is placed at the final valid
token of response $a_k$, with zero reward at earlier generated positions:
\begin{equation}
 r^{\mathrm{turn}}_k=r_k,
 \qquad
 r^{\mathrm{tok}}_{k,j}=
 \begin{cases}
  0, & j<L_k,\\
  r_k, & j=L_k
 \end{cases}.
 \label{eq:reward-layout}
\end{equation}
The environment-specific components of $r_k$ are summarized in
Appendix~\ref{app:task-interaction-rewards}.
For each granularity $b\in\{\mathrm{turn},\mathrm{tok}\}$, let $i$ denote a turn $k$
or a valid generated-token position $(k,j)$, respectively, and let
$\operatorname{succ}_b(i)$ denote its next position in the corresponding
sequence. Figure~\ref{fig:method-overview}(c) shows turn-level and token-level
views of the same trajectory. The turn-level trajectory advances to turn $k+1$;
the token-level trajectory advances to the next generated token within the response or,
after its final token, to the first generated token of the next response.

Each granularity uses its own discount factor $\gamma_b\in[0,1]$ and
GAE trace parameter $\lambda_b\in[0,1]$, which control reward discounting
and the decay of future temporal-difference residuals, respectively.
Let $V_{\mathrm{old}}^b$ denote the critic's value estimates before the
current optimization phase. These estimates are used to compute the
temporal-difference residual $\delta_i^b$ and advantage estimate $A_i^b$:
\begin{equation}
  \begin{aligned}
    \delta_i^b
      &=r_i^b+\gamma_b V_{\mathrm{old}}^b
        \!\left(\operatorname{succ}_b(i)\right)-V_{\mathrm{old}}^b(i),\\
    A_i^b
      &=\delta_i^b+\gamma_b\lambda_b
        A_{\operatorname{succ}_b(i)}^b.
  \end{aligned}
  \label{eq:branch-gae}
\end{equation}
The observation between two responses enters the next context but is not a
policy position, so the token-level recursion skips its tokens.  At the final
collected position of each rollout, we set the value and advantage of the
implicit next position to zero. This position is not an additional
critic-supervision target.

Directly adding turn- and token-level advantages would combine their
response-wide offsets as well as the token-level variation. We instead use
the turn-level advantage to determine the response-wide level of credit and
the token-level residual to adjust relative credit within the response.
To prevent the token component from shifting the response-wide mean, we
center its advantages over the valid tokens of each response:
\begin{equation}
  \overline{\Atok}_k
    =\frac{1}{L_k}\sum_{j=1}^{L_k}\Atok_{k,j},
  \qquad
  \widetilde{\Atok}_{k,j}
    =\Atok_{k,j}-\overline{\Atok}_k.
  \label{eq:token-centering}
\end{equation}
With $\alpha$ controlling the residual's relative scale, the final VETTA
advantage is
\begin{equation}
  A^{\mathrm{VETTA}}_{k,j}
    =\Aturn_k+\alpha\widetilde{\Atok}_{k,j}.
  \label{eq:vetta-advantage}
\end{equation}
The centered term $\widetilde{\Atok}_{k,j}$ is the within-response residual.
Appendix~\ref{app:identity} shows that, among vectors with response-wise mean
$\Aturn_k$, the fused advantage vector is closest in squared Euclidean distance
to the scaled token advantages $\alpha(\Atok_{k,j})_{j=1}^{L_k}$.

\subsection{Critic and actor updates}
\label{sec:updates}

At each valid position $i$ of granularity $b$, the corresponding value head
is trained with the return target
\begin{equation}
  \widehat G^b(i)
  =\operatorname{sg}\!\left(V_{\mathrm{old}}^b(i)+A_i^b\right).
  \label{eq:critic-target}
\end{equation}
Here $\operatorname{sg}$ stops gradients through the target. Advantages and
return targets are computed before optimization and held fixed throughout
the corresponding updates. Each value target uses its own granularity's
advantage, before centering or fusion. The critic uses
standard PPO clipped value regression, averaging turn and token losses over
their respective valid positions before summing them.  The explicit loss is
given in Appendix~\ref{app:critic-loss}.

The actor retains token-wise PPO ratios and clipping.  Let
$\theta_{\mathrm{old}}$ denote the policy parameters used to collect the
trajectory.  For each valid generated token $y_{k,j}$, let
$\rho_{k,j}(\theta)$ be the ratio of its probability under $\pi_\theta$
to that under $\pi_{\theta_{\mathrm{old}}}$, with both policies conditioned on
$(H_k,y_{k,<j})$.  The surrogate loss uses the composed advantage:
\begin{equation}
    \mathcal L_{\mathrm{PPO}}
      =-\mathbb E_{k,j}\!\left[
        \min\!\left(
          \rho_{k,j}A^{\mathrm{VETTA}}_{k,j},
          \operatorname{clip}(\rho_{k,j},1-\epsilon,1+\epsilon)
          A^{\mathrm{VETTA}}_{k,j}
        \right)
      \right].
  \label{eq:actor-update}
\end{equation}
Here $\epsilon>0$ is the PPO ratio-clipping radius and the expectation is
over valid generated positions in the rollout batch. The actor implementation's
dual-clipping and KL settings are described in
Appendix~\ref{app:actor-options}.

\section{Experiments}
\label{sec:setup}

We evaluate \method across four multi-turn agent benchmarks and organize the
experiments around four research questions: \textbf{RQ1:} How does \method
compare with prompting and RL methods? \textbf{RQ2:} How does composing
turn- and token-level credit affect performance? \textbf{RQ3:} How do
backbone sharing and value-head structure affect performance?
\textbf{RQ4:} How does critic backbone depth affect performance and
critic-side computation?

\subsection{Experiment Setup}

\paragraph{Benchmarks.}
We evaluate \method on ALFWorld, WebShop, Sokoban, and ScienceWorld. ALFWorld
\citep{shridhar2021alfworld} is a text-based embodied benchmark comprising six
categories of household tasks; we report success rates for each category and
overall.  WebShop
\citep{yao2022webshop} asks agents to search and navigate a simulated shopping
website, and we report task score and success rate.
Sokoban~\citep{gymsokoban} is a visual box-pushing task; we report success
rates under sparse and dense rewards. ScienceWorld~\citep{wang2022scienceworld}
is a text-based environment for scientific tasks; we report overall success
rate and task score.
Evaluation-set sizes and aggregation details appear in
Appendix~\ref{app:eval-sets}.

\paragraph{Baselines.}
On ALFWorld and WebShop, we first compare VETTA with prompting methods:
Qwen2.5-Instruct without task-specific RL training,
ReAct~\citep{yao2022react}, which interleaves
reasoning and actions, and Reflexion~\citep{shinn2023reflexion}, which uses
verbal feedback from previous attempts. These methods do not update the
policy with RL. We then group the RL baselines by whether they train a
parametric critic. The critic-free baselines RLOO~\citep{kool2019buy4,ahmadian2024backtobasics} and
GRPO~\citep{shao2024deepseekmath} estimate relative credit from groups of
sampled trajectories. GiGPO~\citep{feng2025gigpo} further combines
episode-relative credit with action-level comparisons at repeated
environment states.
Among critic-based baselines, PPO~\citep{schulman2017ppo,feng2025gigpo}
copies the episode reward to the final generated token of each response and
computes GAE independently within each response. To better compare PPO with
\method, we implement PPO*, which instead computes GAE across turns over
generated tokens, skipping observation tokens. Both PPO variants use
token-level values, probability ratios, and clipping. Turn-PPO~\citep{li2025turnppo}
instead estimates turn-level values and advantages and applies PPO ratios and
clipping to complete responses.
We train and evaluate GRPO, GiGPO, PPO*, Turn-PPO, and VETTA on Sokoban and
ScienceWorld under the settings below.

\begin{table}[t]
\caption{Performance on ALFWorld and WebShop. Locally evaluated RL methods are reported over three decoding seeds. For ALFWorld, we report success rates (\%) for six
task categories and overall; for WebShop, task score and success rate (\%).
$\dagger$ indicates results reported in
GiGPO~\citep{feng2025gigpo}. The best values are in bold, and the second-best
values are underlined.}
\label{tab:main-agent}
\centering
\scriptsize
\setlength{\tabcolsep}{1.7pt}
\resizebox{\linewidth}{!}{%
\begin{tabular}{llccccccc|cc}
\toprule
\multirow{2}{*}{Type} &
\multirow{2}{*}{Method} &
\multicolumn{7}{c|}{\textbf{ALFWorld}} &
\multicolumn{2}{c}{\textbf{WebShop}} \\
& & Pick & Look & Clean & Heat & Cool & Pick2 & All & Score & Succ. \\
\midrule
\multicolumn{11}{l}{\textit{Qwen2.5-1.5B-Instruct}} \\
Prompting & Qwen2.5$^\dagger$ & 5.9 & 5.5 & 3.3 & 9.7 & 4.2 & 0.0 & 4.1 & 23.1 & 5.2 \\
Prompting & ReAct$^\dagger$ & 17.4 & 20.5 & 15.7 & 6.2 & 7.7 & 2.0 & 12.8 & 40.1 & 11.3 \\
Prompting & Reflexion$^\dagger$ & 35.3 & 22.2 & 21.7 & 13.6 & 19.4 & 3.7 & 21.8 & 55.8 & 21.9 \\
Critic-free & RLOO$^\dagger$ & 88.3$_{\pm3.0}$ & 52.8$_{\pm8.6}$ & 71.0$_{\pm5.9}$ & 62.8$_{\pm8.7}$ & 66.4$_{\pm5.5}$ & 56.9$_{\pm4.7}$ & 69.7$_{\pm2.5}$ & 73.9$_{\pm5.6}$ & 52.1$_{\pm6.7}$ \\
Critic-free & GRPO$^\dagger$ & 85.3$_{\pm1.5}$ & 53.7$_{\pm8.0}$ & 84.5$_{\pm6.8}$ & 78.2$_{\pm7.9}$ & 59.7$_{\pm5.0}$ & 53.5$_{\pm5.6}$ & 72.8$_{\pm3.6}$ & 75.8$_{\pm3.5}$ & 56.8$_{\pm3.8}$ \\
Critic-free & GiGPO$^\dagger$ & \underline{96.0}$_{\pm1.4}$ & \underline{76.5}$_{\pm3.9}$ & \underline{91.8}$_{\pm5.5}$ & \textbf{91.3}$_{\pm6.3}$ & 71.7$_{\pm8.4}$ & \underline{79.5}$_{\pm7.7}$ & \underline{86.1}$_{\pm4.7}$ & \underline{83.5}$_{\pm1.8}$ & \underline{67.4}$_{\pm4.5}$ \\
Critic-based & PPO$^\dagger$ & 64.8$_{\pm3.5}$ & 40.5$_{\pm6.9}$ & 57.1$_{\pm4.9}$ & 60.6$_{\pm6.6}$ & 46.4$_{\pm4.0}$ & 47.4$_{\pm1.9}$ & 54.4$_{\pm3.1}$ & 73.8$_{\pm3.0}$ & 51.5$_{\pm2.9}$ \\
Critic-based & PPO* & 75.2$_{\pm4.4}$ & 48.7$_{\pm4.4}$ & 72.0$_{\pm6.7}$ & 72.5$_{\pm6.8}$ & 72.7$_{\pm7.9}$ & 33.3$_{\pm2.6}$ & 64.8$_{\pm3.7}$ & 70.8$_{\pm1.2}$ & 56.6$_{\pm1.2}$ \\
Critic-based & Turn-PPO & 86.7$_{\pm4.4}$ & 71.8$_{\pm4.4}$ & 70.4$_{\pm6.4}$ & 66.7$_{\pm9.6}$ & \underline{81.3}$_{\pm6.1}$ & 72.2$_{\pm6.4}$ & 76.4$_{\pm2.5}$ & 69.2$_{\pm4.2}$ & 52.2$_{\pm2.9}$ \\
\rowcolor{gray!12}
\cellcolor{white}Critic-based & \textbf{VETTA} & \textbf{97.1}$_{\pm2.9}$ & \textbf{79.5}$_{\pm4.4}$ & \textbf{95.1}$_{\pm4.3}$ & \underline{85.4}$_{\pm9.5}$ & \textbf{88.0}$_{\pm4.0}$ & \textbf{95.8}$_{\pm0.0}$ & \textbf{91.9}$_{\pm1.1}$ & \textbf{83.9}$_{\pm0.6}$ & \textbf{73.8}$_{\pm1.1}$ \\
\midrule
\multicolumn{11}{l}{\textit{Qwen2.5-7B-Instruct}} \\
Prompting & Qwen2.5$^\dagger$ & 33.4 & 21.6 & 19.3 & 6.9 & 2.8 & 3.2 & 14.8 & 26.4 & 7.8 \\
Prompting & ReAct$^\dagger$ & 48.5 & 35.4 & 34.3 & 13.2 & 18.2 & 17.6 & 31.2 & 46.2 & 19.5 \\
Prompting & Reflexion$^\dagger$ & 62.0 & 41.6 & 44.9 & 30.9 & 36.3 & 23.8 & 42.7 & 58.1 & 28.8 \\
Critic-free & RLOO$^\dagger$ & 87.6$_{\pm4.3}$ & 78.2$_{\pm8.3}$ & 87.3$_{\pm5.8}$ & 81.3$_{\pm7.6}$ & 71.9$_{\pm5.2}$ & 48.9$_{\pm8.4}$ & 75.5$_{\pm4.6}$ & 80.3$_{\pm3.2}$ & 65.7$_{\pm4.0}$ \\
Critic-free & GRPO$^\dagger$ & 90.8$_{\pm5.1}$ & 66.1$_{\pm6.7}$ & 89.3$_{\pm5.4}$ & 74.7$_{\pm6.9}$ & 72.5$_{\pm5.4}$ & 64.7$_{\pm7.3}$ & 77.6$_{\pm5.2}$ & 79.3$_{\pm2.8}$ & 66.1$_{\pm3.7}$ \\
Critic-free & GiGPO$^\dagger$ & 91.8$_{\pm5.4}$ & \textbf{88.6}$_{\pm6.3}$ & \underline{95.9}$_{\pm3.2}$ & \underline{90.2}$_{\pm2.6}$ & \underline{86.5}$_{\pm5.5}$ & 85.2$_{\pm7.5}$ & 90.2$_{\pm2.3}$ & \underline{86.2}$_{\pm2.6}$ & \underline{75.2}$_{\pm3.8}$ \\
Critic-based & PPO$^\dagger$ & 92.3$_{\pm4.0}$ & 64.0$_{\pm8.4}$ & 92.5$_{\pm2.4}$ & 89.5$_{\pm7.0}$ & 80.3$_{\pm2.0}$ & 68.8$_{\pm8.3}$ & 80.4$_{\pm2.7}$ & 81.4$_{\pm3.1}$ & 68.7$_{\pm5.1}$ \\
Critic-based & PPO* & 92.1$_{\pm1.0}$ & 85.7$_{\pm7.6}$ & 77.4$_{\pm8.3}$ & 72.2$_{\pm2.4}$ & 55.6$_{\pm9.6}$ & 57.6$_{\pm11.1}$ & 72.6$_{\pm2.1}$ & 84.5$_{\pm0.1}$ & 74.8$_{\pm0.4}$ \\
Critic-based & Turn-PPO & \textbf{100.0}$_{\pm0.0}$ & 82.9$_{\pm8.9}$ & \textbf{100.0}$_{\pm0.0}$ & \textbf{100.0}$_{\pm0.0}$ & 84.5$_{\pm2.3}$ & \underline{92.4}$_{\pm4.2}$ & \underline{94.3}$_{\pm1.5}$ & 75.3$_{\pm0.8}$ & 54.6$_{\pm1.9}$ \\
\rowcolor{gray!12}
\cellcolor{white}Critic-based & \textbf{VETTA} & \underline{93.6}$_{\pm2.6}$ & \underline{86.7}$_{\pm5.8}$ & \textbf{100.0}$_{\pm0.0}$ & 90.0$_{\pm0.0}$ & \textbf{93.9}$_{\pm3.0}$ & \textbf{100.0}$_{\pm0.0}$ & \textbf{95.5}$_{\pm0.4}$ & \textbf{86.5}$_{\pm0.7}$ & \textbf{76.0}$_{\pm0.6}$ \\
\bottomrule
\end{tabular}}
\end{table}

\paragraph{Implementation details.}
We use Qwen2.5-1.5B/7B-Instruct~\citep{qwen2025qwen25} as our base
models for ALFWorld and WebShop. Within each model-size comparison, methods
use the same base actor model. Sokoban uses
Qwen2.5-VL-3B-Instruct~\citep{qwen2025vl3b}, and ScienceWorld uses
Qwen2.5-7B-Instruct. PPO* and Turn-PPO retain the full critic depth: 36
decoder blocks on Sokoban and 28 on the text benchmarks. \method retains
two critic blocks on all four benchmarks. We set
$\gamma_{\mathrm{tok}}=\lambda_{\mathrm{tok}}=1$ and
$\gamma_{\mathrm{turn}}=\lambda_{\mathrm{turn}}=0.95$ for VETTA on all four benchmarks.
Unless otherwise specified, the token-residual scale is $\alpha=1$ for
ALFWorld and Sokoban, $\alpha=3$ for WebShop, and $\alpha=0.25$ for
ScienceWorld. Training runs for 150 updates on each benchmark.
For our local evaluations on ALFWorld, WebShop, and Sokoban, we report means
and sample standard deviations over three decoding seeds. ScienceWorld is
evaluated once using greedy decoding. Additional optimization settings are
given in Table~\ref{tab:training-config}, with evaluation details in
Appendix~\ref{app:eval-sets}.

\FloatBarrier
\subsection{Experimental Results (RQ1)}
\label{sec:results}
\label{sec:overall-results}

Tables~\ref{tab:main-agent} and~\ref{tab:sokoban-scienceworld} show that
\method achieves strong overall success across ALFWorld, WebShop, Sokoban,
and ScienceWorld. On ALFWorld and WebShop, ReAct and Reflexion improve over the
Qwen2.5-Instruct prompting baseline, but their overall success rates remain
well below those of the RL-trained methods. This establishes the value of
learning from task feedback in both interactive settings.

Among critic-free RL methods, RLOO and GRPO substantially improve over
prompting, although their advantage over PPO varies across benchmarks and
model sizes. GiGPO further assigns action-level relative credit using
repeated environment states and achieves the highest overall success rates
among the critic-free baselines on ALFWorld and WebShop.

On ALFWorld and WebShop, the critic-based baselines reveal different
strengths. PPO* propagates token-level GAE across turns while retaining
token-wise PPO updates, whereas Turn-PPO estimates advantages for complete
responses. At 1.5B, Turn-PPO is stronger on ALFWorld, while PPO* performs
slightly better on WebShop; both remain below \method. This contrast becomes
more pronounced at 7B: Turn-PPO reaches 94.3\% overall success on ALFWorld,
whereas PPO* reaches 74.8\% success on WebShop. Neither single-granularity
baseline maintains the same strength across both benchmarks.

\method uses turn-level advantages to provide response-level credit across interactions
and token-level residuals to distinguish decisions within each response. At
1.5B, it reaches 91.9\% success on ALFWorld and 73.8\% on WebShop, exceeding
the strongest alternative in each benchmark by 5.8 and 6.4 percentage
points, respectively. At 7B, it retains the highest mean overall success
rate among both critic-based and critic-free baselines on both benchmarks,
reaching 95.5\% and 76.0\%;
it also achieves the highest WebShop task score at both model sizes.

\begin{table}[t]
\centering
\small
\setlength{\tabcolsep}{4pt}
\caption{Performance on Sokoban and ScienceWorld. Sokoban results are
reported over three decoding seeds. For Sokoban, we train
Qwen2.5-VL-3B-Instruct under sparse and dense rewards and report success
rates (\%); for ScienceWorld, we use Qwen2.5-7B-Instruct and report task
score and success rate (\%). The best values are in bold, and the
second-best values are underlined.}
\label{tab:sokoban-scienceworld}
\begin{tabular}{llcc|cc}
\toprule
\multirow{2}{*}{Type} & \multirow{2}{*}{Method}
& \textbf{Sokoban (Sparse)} & \textbf{Sokoban (Dense)}
& \multicolumn{2}{c}{\textbf{ScienceWorld}} \\
& & Succ. (\%) & Succ. (\%) & Score & Succ. (\%) \\
\midrule
Critic-free & GRPO & 28.65$_{\pm 2.96}$ & 60.94$_{\pm 2.34}$ & 20.37 & 38.10 \\
Critic-free & GiGPO & 63.80$_{\pm 2.51}$ & 47.66$_{\pm 2.07}$ & 18.84 & 40.24 \\
Critic-based & PPO* & 35.42$_{\pm 3.52}$ & 29.17$_{\pm 3.69}$ & 19.19 & 40.63 \\
Critic-based & Turn-PPO & \underline{78.13}$_{\pm 1.56}$ & \underline{78.65}$_{\pm 1.97}$ & \underline{24.34} & \underline{50.69} \\
\rowcolor{gray!12}
\cellcolor{white}Critic-based & \textbf{VETTA} & \textbf{84.90}$_{\pm 1.19}$ & \textbf{80.73}$_{\pm 1.19}$ & \textbf{25.89} & \textbf{52.17} \\
\bottomrule
\end{tabular}
\end{table}

On Sokoban, \method achieves the highest mean success rate under both
sparse and dense rewards, reaching 84.90\% and 80.73\%, respectively.
Turn-PPO is the strongest baseline in both settings, while the relative
performance of GRPO and GiGPO changes with the reward setting. \method's
advantage in both settings indicates that its effectiveness is not
restricted to sparse task rewards.

On ScienceWorld, \method reaches 52.17\% overall success and a task
score of 25.89, compared with 50.69\% and 24.34 for Turn-PPO. It also
improves overall success over PPO* by 11.54 percentage points. The gains
in both overall completion and task score support its effectiveness on
scientific interaction tasks.

\FloatBarrier
\subsection{Further Analysis}
\label{sec:further-analysis}
We further examine how credit composition, critic sharing, and backbone depth affect
\method's performance.

\paragraph{Credit composition (RQ2).}
To assess the contribution of each credit level and within-response
centering, we compare actor-credit rules on WebShop using
Qwen2.5-1.5B-Instruct.

We keep the shared two-block critic and the token-wise PPO ratios and
clipping in Eq.~\eqref{eq:actor-update} fixed. Rows C1--C4 train both value
heads and change only the advantage given to the actor. Direct addition (C3)
and residual composition (C4) use $\alpha=3$, so their comparison isolates
within-response centering. In C5 and C6, the actor uses one credit branch;
the corresponding value loss trains its output head and the critic's
Transformer backbone.

\begin{embeddedpair}
\begin{minipage}[t]{0.57\linewidth}
\vspace{0pt}
\setlength{\abovecaptionskip}{0pt}
\captionof{table}{Analysis of credit composition on WebShop.}
\label{tab:credit-composition}
\centering
\small
\setlength{\tabcolsep}{5pt}
\begin{tabular}{@{}clcrr@{}}
\toprule
ID & Actor credit & \shortstack{Critic\\supervision} & \shortstack{Success\\rate (\%)} & \shortstack{Task\\score} \\
\midrule
C1 & $\Atok$ & Turn + token & 67.4 & \textbf{84.9} \\
C2 & $\Aturn$ & Turn + token & 59.8 & 74.5 \\
C3 & $\Aturn+\alpha\Atok$ & Turn + token & 55.0 & 79.0 \\
C4 & $\Aturn+\alpha\widetilde{\Atok}$ & Turn + token & \textbf{72.8} & 83.6 \\
\midrule
C5 & $\Atok$ & Token & 44.8 & 56.8 \\
C6 & $\Aturn$ & Turn & 68.0 & 84.5 \\
\bottomrule
\end{tabular}

\end{minipage}\hfill
\begin{minipage}[t]{0.40\linewidth}
\vspace{0pt}
Among the matched dual-head runs in Table~\ref{tab:credit-composition}, C4
achieves the highest success rate (72.8\%). Its 17.8-point gain over direct
addition C3 (55.0\%) demonstrates that relative token advantages yield
higher success than uncentered token advantages when combined with
turn-level credit. C4 also outperforms token-only actor credit C1 (67.4\%)
and turn-only actor credit C2 (59.8\%).
\end{minipage}
\end{embeddedpair}

C1 nevertheless achieves the highest task score (84.9), but its success
rate (67.4\%) is lower than C4's (72.8\%). Higher task scores therefore do
not necessarily imply more successful completions. In our WebShop setting,
only complete success receives a positive task reward
(Appendix~\ref{app:task-interaction-rewards}).

The single-branch control C5 reaches 44.8\% success, while C6 reaches
68.0\% success and a task score of 84.5, slightly above C4's 83.6. Yet C4
has a higher success rate than both controls. Residual composition achieves
the highest success rate, while the higher task scores of C1 and C6 show
that graded task score and exact completion favor different credit
configurations.

\paragraph{Critic sharing and value heads (RQ3).}
The critic in \method makes two structural choices: the turn and token values
share a backbone but have separate output heads. We examine both choices on
ALFWorld and WebShop using Qwen2.5-1.5B-Instruct. To assess backbone sharing,
we compare two independent two-block critics, each with its own value head,
against one two-block backbone with separate turn- and token-value heads. To
assess the heads, we also include a shared-backbone variant with one output
head trained on both value targets. At the first generated-token position of
each turn, the same prediction serves as both the turn value and the first
token value; later positions provide the remaining token values. All variants
retain turn- and token-level credit and the residual composition rule.
\begin{figure}[!t]
  \centering
  \includegraphics[width=\linewidth]{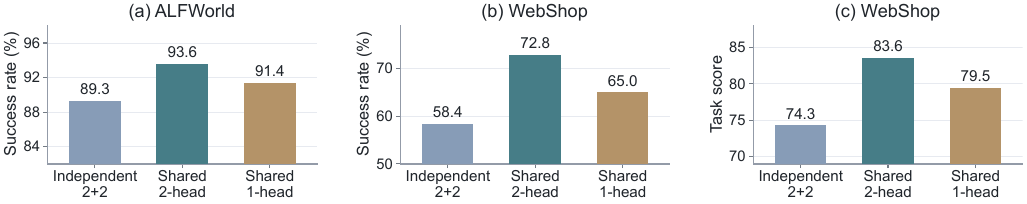}
  \caption{Critic structure: (a) ALFWorld success rate, (b) WebShop success rate, and
  (c) WebShop task score.}
  \label{fig:critic-structure}
\end{figure}

Figure~\ref{fig:critic-structure} shows that the shared two-head critic has
the highest success rate on both benchmarks: 93.6\% on ALFWorld and 72.8\%
on WebShop. The independent critics reach 89.3\% and 58.4\%, while the
shared one-head critic reaches 91.4\% and 65.0\%, respectively. WebShop task
scores follow the same ordering. The shared two-head critic outperforms
both independent critics and the shared one-head variant, demonstrating
the effectiveness of backbone sharing with separate value heads.

\paragraph{Critic depth and computation (RQ4).}
We ask how much critic computation can be saved by retaining only the early
Transformer blocks. On each benchmark, we compare 2-, 8-, and 28-block
critics while keeping both value heads and the actor-credit rule unchanged.

\begin{embeddedpair}
\begin{minipage}[t]{0.55\linewidth}
\vspace{0pt}
Figure~\ref{fig:critic-depth-tradeoff} shows the savings in critic-side
computation, measured over 112 training steps per run
(Appendix~\ref{app:critic-depth-timing}). Compared
with 28 blocks, the two-block critic records 19.7 rather than 152.0 seconds
of median value-inference-plus-update time on ALFWorld, approximately 87\%
less. On WebShop, the corresponding times are 7.9 and 92.2 seconds,
approximately 91\% less. Its median share of training-step time likewise
falls from 25.6\% to 3.8\% on ALFWorld and from 27.8\% to 4.3\% on WebShop;
the eight-block runs lie between these endpoints.

The two-block model saves computation by removing 26 Transformer blocks
from its trainable backbone. Despite this
reduction, Figure~\ref{fig:intro-observations}(b) shows the highest mean
success rate with two blocks on both benchmarks.
\end{minipage}\hfill
\begin{minipage}[t]{0.42\linewidth}
\vspace{0pt}
\centering
\includegraphics[width=\linewidth]{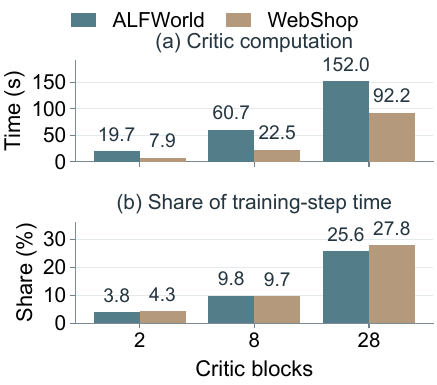}
\captionof{figure}{Critic computation by backbone depth on ALFWorld and WebShop.}
\label{fig:critic-depth-tradeoff}

\end{minipage}
\end{embeddedpair}

The eight-block results
further show that retaining more layers does not necessarily improve success.
On both benchmarks, the two-block critic reduces critic-side time while
achieving higher success rates than the full-depth critic.
\FloatBarrier

\section{Related Work}
\label{sec:related}

In this section, we position \method{} relative to prior work on turn-level credit,
token-level and joint credit, and critic design.

\paragraph{Turn-Level Credit Assignment.}
Turn-level methods assign credit to complete textual actions within a
multi-turn trajectory. Critic-free approaches derive action-level signals
from sampled returns using state- or history-based groups
\citep{feng2025gigpo,he2026hgpo}, or from progress toward successful states in
a constructed transition graph \citep{cheng2026graphgpo}. Some explicitly
estimate values without training a parametric critic: GAGPO constructs
grouped return proxies for temporal-difference (TD) and generalized advantage
estimation (GAE)
\citep{zhu2026gagpo}, while Gated-BEPO estimates node values through Bellman
backups on empirical rollout graphs \citep{yan2026gatedbepo}.
Learned value functions provide another route. ArCHer learns utterance-level
values to guide a lower-level token policy \citep{zhou2024archer}, and
Turn-PPO learns turn-level values and applies PPO ratios and clipping to
complete responses \citep{li2025turnppo}. These approaches distinguish the
contributions of different actions; our focus is on combining such action-level
credit with variation among the token-generation decisions within each action.

\paragraph{Token-Level and Joint Credit Assignment.}
Other work differentiates credit within a response or jointly models multiple
granularities. SAO connects token-level GAE across adjacent responses while
skipping observation tokens in the recursion \citep{hou2026sao}.
FACTOR allocates action credit to executable tokens with nonnegative mean-one
weights \citep{ma2026factor}, while SERL uses post-action teacher feedback to
adjust token updates selectively \citep{li2026serl}.
The closest to our joint estimation is HyGAE, which linearly combines turn-
and token-level advantages and trains a unified value function with mixed
return targets \citep{zhang2026hygae}. \method{} instead retains separate
supervision for the two value heads and adds a centered token-level residual
to the turn-level advantage. Centering makes the residual average to zero
within each response, so the turn advantage determines the mean of the
composed signal, while token-level variation
adjusts credit within that response. This separates the value-learning
targets and the roles of the two advantage components.

\paragraph{Critic Architecture and Efficiency.}
Value-model design also determines how representations are shared and how
much computation value learning requires. HiPER shares a critic backbone
across value heads for subgoal planning and action execution
\citep{peng2026hiper}; \method{} uses shared representations for complete
responses and individual token-generation decisions.
To reduce critic cost, AsyPPO uses ensembles of smaller critics
\citep{liu2025asyppo}, while POISE fits lightweight value probes to the
actor's internal representations \citep{choi2026poise}.
\method{} constructs its shared critic backbone from the early Transformer
blocks of the pretrained checkpoint used to initialize the actor, reducing
backbone depth while retaining both value-estimation objectives.
Complementary work such as VC-PPO and VAPO addresses value-learning failures
and optimization stability
\citep{yuan2025vcppo,yue2025vapo}.

\section{Conclusion}

We introduce \method, a critic-based framework for multi-turn LLM agents
that estimates turn- and token-level values with a shared lightweight
backbone and separate value heads. It combines the turn-level advantage
with a centered token-level residual to guide policy optimization. Across
all four benchmarks, \method outperforms both critic-based and critic-free
baselines in overall success rate. Its strong performance on Sokoban holds under both
sparse and dense rewards. Depth comparisons on ALFWorld and WebShop show that retaining
only early Transformer blocks preserves strong performance while substantially
reducing critic-side time. Together, these results highlight credit granularity
and critic efficiency as complementary design choices.

\section*{AI use statement}

Generative AI tools were used to draft and polish manuscript text, search for
and summarize related literature, provide feedback on experimental comparisons
and the interpretation of results, and assist with debugging and reproduction
checks. They also assisted with preparing figures, tables, and manuscript
scripts. AI tools did not directly operate the experiments or implement the proposed
training method. The authors take responsibility for the final manuscript,
including AI-assisted text, analyses, and artifacts.

\section*{Reproducibility statement}

Section~\ref{sec:method} describes the value estimates, advantage computation,
and optimization objectives. Section~\ref{sec:setup} and
Appendices~\ref{app:config} and~\ref{app:eval-sets} provide the training
and evaluation configurations.

\bibliography{references}
\bibliographystyle{styles/iclr2027/iclr2027_conference}

\appendix
\section{Additional Analyses}

\subsection{Sensitivity to the token-residual scale}
\label{app:alpha-sensitivity}

\begin{figure}[H]
  \centering
  \includegraphics[width=\linewidth]{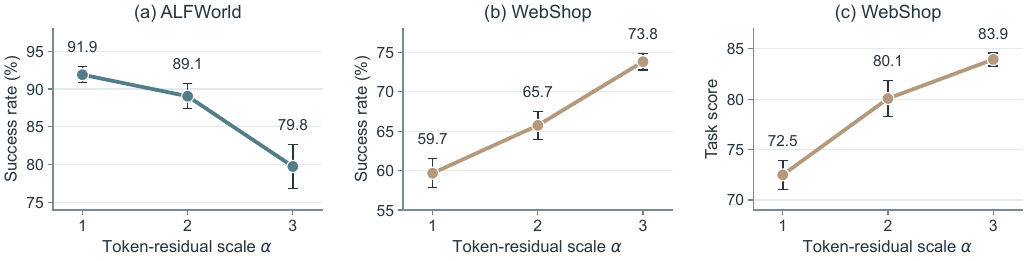}
  \caption{Sensitivity to the token-residual scale $\alpha$ with
  Qwen2.5-1.5B-Instruct. Markers and error bars show the mean and sample
  standard deviation over three decoding seeds.}
  \label{fig:alpha-sensitivity}
\end{figure}

We examine the residual weight $\alpha$ in
Equation~\ref{eq:vetta-advantage} for the 1.5B shared-critic configuration.
Figure~\ref{fig:alpha-sensitivity} reports ALFWorld success rate, WebShop
success rate, and WebShop task score for $\alpha\in\{1,2,3\}$.

On ALFWorld, mean success declines from 91.9\% at $\alpha=1$ to 89.1\% at
$\alpha=2$ and 79.8\% at $\alpha=3$. WebShop shows the opposite trend:
mean success rises from 59.7\% to 65.7\% and 73.8\%, while mean task score
rises from 72.5 to 80.1 and 83.9. These opposing trends suggest a
task-dependent balance: among the tested settings, ALFWorld favors a smaller
token-residual scale, whereas WebShop favors a larger one. One possible
explanation is that ALFWorld's action outcomes provide relatively clear
turn-level progress signals, while WebShop's search and selection actions
may benefit more from distinguishing generated tokens within each response.
Each $\alpha$ corresponds to a separately trained checkpoint, and the error
bars reflect decoding-seed variation rather than independent training runs.
\FloatBarrier

\subsection{Turn and token credit in an ALFWorld trajectory}
\label{app:credit-case}

We illustrate the two levels of credit with a successful ALFWorld trajectory
on the seen task ``look at book under the desklamp'' (latest step-150
checkpoint, environment seed~0, decoding seed~123). The agent moves to
desk~1, picks up book~4, and uses desklamp~1 to complete the task.

\begin{figure}[H]
  \centering
  \includegraphics[width=\linewidth]{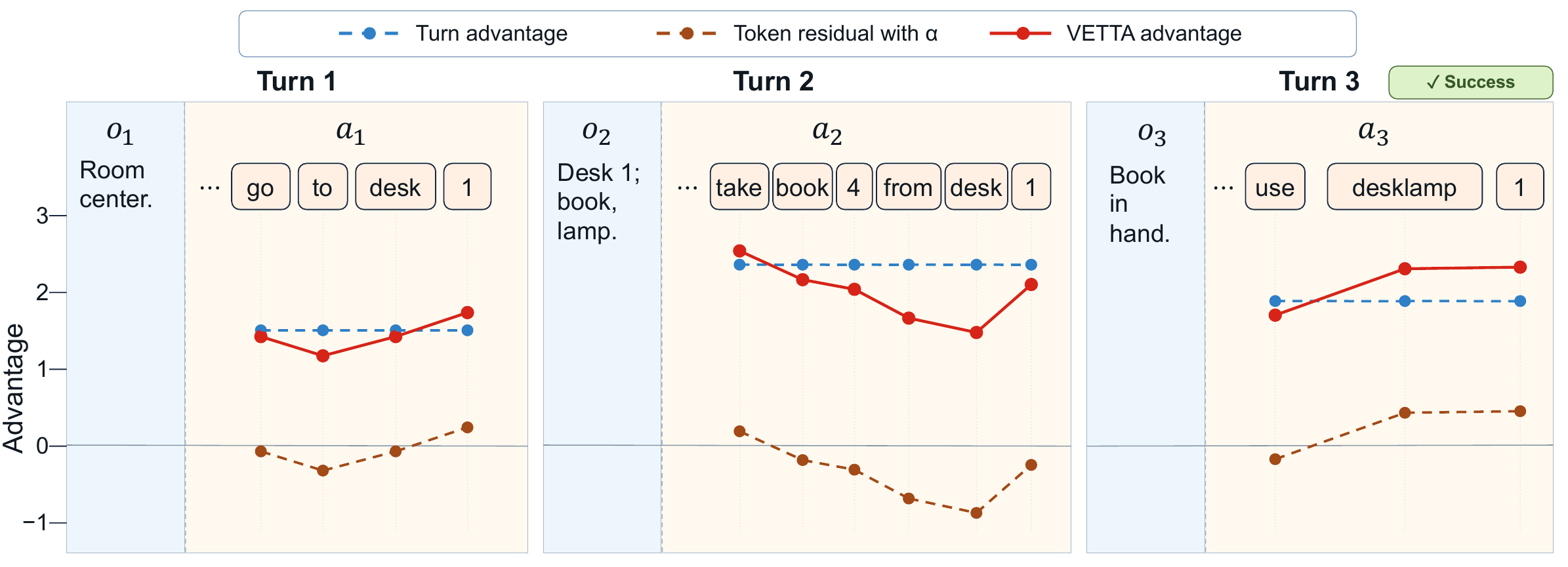}
  \caption{Turn- and token-level credit in a successful ALFWorld trajectory.
  Blue, brown, and red denote $\Aturn_k$, $\alpha\widetilde{\Atok}_{k,j}$,
  and $A^{\mathrm{VETTA}}_{k,j}$. Only parsed actions are shown; thinking text
  is abbreviated. When a plotted word spans multiple generated tokens, its
  values are arithmetic means over those token positions.}
  \label{fig:alfworld-credit-case}
\end{figure}

The three turns receive positive turn advantages (1.494, 2.348, and 1.875),
providing response-level credit. Token residuals distinguish words within
each response: in the pickup action, \texttt{take} receives $+0.180$ and
\texttt{from} receives $-0.695$; in the final action, \texttt{desklamp}
and \texttt{1} receive $+0.421$ and $+0.441$. The latter two words have
fused advantages of 2.296 and 2.316, above their turn-level advantage of 1.875.
The turn advantage sets the credit level for each response, while the token
residual refines relative credit within that response.

The displayed residuals need not average to zero because the figure omits
thinking text. Centering uses all valid generated tokens in the complete
response, including the omitted text, so its full-response residual mean is
zero by construction.

\FloatBarrier

\section{Experimental Details}

\subsection{Task interaction and rewards}
\label{app:task-interaction-rewards}

For \method on all four benchmarks, each turn's task reward and any
applicable action penalty are assigned to the final valid generated token
of that response, following Equation~\ref{eq:reward-layout}.
When enabled, KL regularization is applied separately to the actor loss
rather than added to these rewards.

\paragraph{ALFWorld.}
In the text-based ALFWorld setting~\citep{shridhar2021alfworld}, the agent
receives a household goal and textual observations, and issues commands to
navigate, inspect, and manipulate objects. The environment executes the
command extracted from each generated response and supplies the next
observation. The text-environment wrapper assigns a task reward of 10 upon
success and 0 otherwise. An action flagged as invalid receives an additional
penalty of 0.1.

\paragraph{WebShop.}
The agent receives a shopping instruction and interacts with the simulated
website through search and click actions~\citep{yao2022webshop}. Observations
describe the current search results or product page. The wrapper retains the
environment's continuous task score for evaluation, but uses a task reward
of 10 only when the purchase ends the episode with a task score of 1; other
outcomes receive 0. Invalid actions incur an additional penalty of 0.1.
Throughout the paper, reported WebShop task scores are multiplied by 100
and lie on a 0--100 scale. This presentation scale does not change the
environment's original score or the training reward.

\paragraph{Sokoban.}
The agent observes a rendered $6\times6$ board with one box and generates
commands to push it onto the target. We train separate policies under sparse
and dense rewards. Sparse rewards assign 10 upon task completion and 0
otherwise. Dense rewards include a step cost of $-0.1$, $+1$ when a box
enters a target, $-1$ when it leaves, and an additional completion reward of
10. The inherited invalid-action penalty of 0.1 is retained in both settings.

\paragraph{ScienceWorld.}
We use ScienceWorld version 1.2.3 with its 30 task types and official
variation splits, without environment simplifications. The agent receives
textual observations and issues commands to complete scientific tasks.
Training uses a task reward of 10 for success and 0 otherwise, with a format
penalty of 0.01. We separately retain the simulator's final task score
for evaluation; this score is not the sparse training reward and can be
negative for unsuccessful episodes.

\subsection{Training settings}
\label{app:config}

Table~\ref{tab:training-config} summarizes the training hyperparameters for
\method on all four benchmarks, including the rollout budget per update,
maximum number of turns, PPO minibatch size, learning rates, KL coefficient,
discount and GAE parameters, and total number of training updates.

\begin{table}[!t]
\caption{Training settings for VETTA on the four benchmarks.}
\label{tab:training-config}
\centering
\small
\setlength{\tabcolsep}{3.5pt}
\begin{tabularx}{\linewidth}{@{}Xrrrr@{}}
\toprule
Setting & ALFWorld & WebShop & Sokoban & ScienceWorld \\
\midrule
Retained critic blocks & 2 & 2 & 2 & 2 \\
Tasks $\times$ rollouts/update & $128\times1$ & $128\times1$ & $128\times1$ & $128\times1$ \\
Maximum turns & 50 & 15 & 15 & 50 \\
PPO minibatch & 256 & 64 & 64 & 256 \\
Actor LR & $10^{-6}$ & $10^{-6}$ & $10^{-6}$ & $5\times10^{-7}$ \\
Critic LR & $10^{-5}$ & $10^{-5}$ & $10^{-5}$ & $10^{-5}$ \\
KL loss coefficient & 0.01 & 0.01 & 0.01 & 0 \\
Token $\gamma/\lambda$ & $1/1$ & $1/1$ & $1/1$ & $1/1$ \\
Turn $\gamma/\lambda$ & $0.95/0.95$ & $0.95/0.95$ & $0.95/0.95$ & $0.95/0.95$ \\
Token-residual scale $\alpha$ & 1 & 3 & 1 & 0.25 \\
Update budget & 150 & 150 & 150 & 150 \\
\bottomrule
\end{tabularx}
\end{table}

Within each Sokoban or ScienceWorld reward setting, all methods collect 128
trajectories per update. GRPO and GiGPO use 16 task roots with eight rollouts
per root; PPO*, Turn-PPO, and VETTA use 128 roots with one rollout each.
Thus the total trajectory budget is equal, while task grouping differs.
On Sokoban, the group-method roots and dense PPO* roots are sampled without
replacement within an update; the VETTA and Turn-PPO roots use the inherited
sampling with replacement.

\subsection{Evaluation sets and aggregation}
\label{app:eval-sets}

Across all four benchmarks, our local evaluations use the latest checkpoint
after 150 training updates. Results reported from GiGPO retain the source
paper's evaluation protocol.

The reported ALFWorld results use the fixed 140-task \texttt{valid\_seen}
split. The WebShop endpoint uses a fixed pool of 500 tasks.
Sokoban uses a fixed ordered set of 128 boards in both reward settings.
ScienceWorld uses all 1,819 official test variations from the 30 task types;
its train and development splits contain 3,592 and 1,796 variations.

Figures~\ref{fig:intro-observations} and~\ref{fig:alpha-sensitivity}, together
with our locally evaluated three-seed results in Table~\ref{tab:main-agent},
use decoding seeds 123, 456, and 789. We evaluate each fixed checkpoint with
these three seeds and report the arithmetic mean and sample standard
deviation across decoding runs. The ALFWorld evaluations use environment
seed 0, temperature 0.4, and sampling.

The credit-composition results in Table~\ref{tab:credit-composition} and the
critic-structure results in Figure~\ref{fig:critic-structure} instead use a
single decoding seed, 0. Consequently,
their success rates can differ from the three-seed averages in
Table~\ref{tab:main-agent}, even for the same configuration. The
depth-performance results in Figure~\ref{fig:intro-observations}(b) use the
three decoding seeds specified above; the critic-time measurements in
Figure~\ref{fig:critic-depth-tradeoff} come from training logs and do not
involve decoding seeds.

All Sokoban methods in Table~\ref{tab:sokoban-scienceworld} are evaluated
over three decoding seeds (123, 456, and 789) under both sparse and dense
reward settings, with temperature 0.4 and at most 15 turns. Each
seed evaluates the same checkpoint on the same 128 boards.
Values are means and sample standard deviations over these three decoding
seeds. Training and evaluation use disjoint sets of initial-board generation
seed IDs, separate from the decoding seeds above.

ScienceWorld results in Table~\ref{tab:sokoban-scienceworld} use one greedy
evaluation of each checkpoint, with engine seed 42, at most
50 turns, history length two, and prompt/response limits of 4,096/512 tokens.
All methods evaluate the same task/variation IDs. Overall success rate is
the fraction of the 1,819 instances completed successfully; task
score is the arithmetic mean of their final simulator scores.

\subsection{Critic-depth timing}
\label{app:critic-depth-timing}

All six depth runs train for 150 updates. For timing, we retain steps 11--149
whose indices are not divisible by five, excluding the first ten training
steps and scheduled validation or checkpoint steps. This gives 112 steps
per run. We add value-inference and critic-update times within each step,
then take the median. The reported time share is the median of the
per-step ratio of this sum to total step time.

Figure~\ref{fig:critic-depth-tradeoff} shows the compact depth comparison.
Table~\ref{tab:critic-depth-timing-detail} reports the underlying stage
medians. These are within-run measurements of critic computation, not an
end-to-end speedup against a critic-free optimizer. ALFWorld's two- and
28-block runs used the same host, as did WebShop's eight- and 28-block runs;
the other depth pairs span hosts.
\begin{table}[!t]
\caption{Critic timing by backbone depth. All entries are medians over 112
clean training steps. Local time is value inference plus critic update within
each step; its share is computed against that step's total time.}
\label{tab:critic-depth-timing-detail}
\centering
\small
\setlength{\tabcolsep}{3pt}
\begin{tabular}{@{}lrrrrr@{}}
\toprule
Benchmark & Blocks & Value (s) & Update (s) & Local (s) & Share (\%) \\
\midrule
ALFWorld & 2  & 3.116 & 16.538 & 19.696 & 3.845 \\
ALFWorld & 8  & 10.585 & 50.460 & 60.737 & 9.763 \\
ALFWorld & 28 & 26.700 & 125.500 & 151.975 & 25.634 \\
\midrule
WebShop & 2  & 1.480 & 6.546 & 7.917 & 4.321 \\
WebShop & 8  & 4.297 & 18.212 & 22.455 & 9.661 \\
WebShop & 28 & 17.579 & 74.459 & 92.224 & 27.783 \\
\bottomrule
\end{tabular}
\end{table}

\FloatBarrier

\section{Method Details and Derivations}

\subsection{Value-position alignment}
\label{app:alignment}

For a response of length $L$, the implementation slices critic logits at
positions \texttt{[-L-1:-1]}.  The first aligned output therefore gives the
value immediately before the first generated token, and the $j$th aligned
output precedes generated token $y_j$.  The complete prompt and response are
encoded in one causal forward pass, but causal attention prevents future
response tokens from changing an
earlier aligned value.  The turn target is supervised only at the first aligned
position, whereas token targets use every valid response position.

\subsection{Valid generated-token positions}
\label{app:valid-tokens}
The rollout stores each turn as a context--response pair. The response mask
includes generated positions through the first end-of-sequence token,
including that token when present, and excludes subsequent padding. If no
end-of-sequence token is emitted before the generation limit, the retained
generated positions remain valid. Generated reasoning text and formatting
tokens are not separately removed. Instructions, previous interaction
history, and environment observations enter
the prompt and are excluded from the current response mask. This mask selects
the token-level value targets, advantage-centering positions, and actor-loss
positions. Token-level GAE connects valid generated positions in temporal
order across responses of the same trajectory; each turn contributes only
one turn-value target at its first aligned position.

\subsection{Critic regression objective}
\label{app:critic-loss}

For granularity $b\in\{\mathrm{turn},\mathrm{tok}\}$ and valid position $i$,
let $\widehat G^b(i)$ be the fixed target in Equation~\ref{eq:critic-target},
$\psi$ the trainable critic parameters, and $\epsilon_v>0$ the
value-clipping radius. The clipped value loss is
\begin{equation}
  \ell^b(i)
  =\max\!\left\{
    \big(V^b_\psi(i)-\widehat G^b(i)\big)^2,
    \left[
      \operatorname{clip}\!\left(
        V^b_\psi(i),
        V^b_{\mathrm{old}}(i)-\epsilon_v,
        V^b_{\mathrm{old}}(i)+\epsilon_v
      \right)-\widehat G^b(i)
    \right]^2
  \right\}.
  \label{eq:clipped-value-loss}
\end{equation}
The two heads are reduced over their own valid positions before their losses
are summed, without dividing the sparse turn loss by the number of tokens:
\begin{equation}
  \mathcal L_{\mathrm{critic}}
  =\operatorname{Mean}_{k=1,\ldots,K}
      [\ell^{\mathrm{turn}}(k)]
   +\operatorname{Mean}_{(k,j):\,1\le k\le K,\,1\le j\le L_k}
      [\ell^{\mathrm{tok}}(k,j)].
  \label{eq:critic-loss}
\end{equation}

\subsection{Response-wise properties of advantage fusion}
\label{app:identity}

Write $D_{k,j}=\widetilde{\Atok}_{k,j}$ for the centered token residual.
For a response with $L_k$ valid generated tokens and fixed $\alpha$, consider
the token-wise vector $\mathbf z=(z_1,\ldots,z_{L_k})$. Among vectors whose
response-wise mean equals the turn advantage, the one closest to the scaled
token advantages solves
\begin{equation}
  \begin{aligned}
    \min_{\mathbf z\in\mathbb R^{L_k}}\quad
      &\frac{1}{2}\sum_{j=1}^{L_k}
        \bigl(z_j-\alpha\Atok_{k,j}\bigr)^2\\
    \text{subject to}\quad
      &\frac{1}{L_k}\sum_{j=1}^{L_k}z_j=\Aturn_k.
  \end{aligned}
  \label{eq:advantage-projection}
\end{equation}
With a Lagrange multiplier $\eta$ for the equivalent constraint
$\sum_j z_j=L_k\Aturn_k$, stationarity gives
$z_j=\alpha\Atok_{k,j}-\eta$.
The constraint yields $\eta=\alpha\overline{\Atok}_k-\Aturn_k$;
thus $z_j=A^{\mathrm{VETTA}}_{k,j}$.

The squared-distance objective is strictly convex, so this solution is unique.
It follows that
\begin{align*}
 \sum_{j=1}^{L_k}D_{k,j}
 &=\sum_{j=1}^{L_k}\left(\Atok_{k,j}-\overline{\Atok}_k\right)=0,\\
 \frac{1}{L_k}\sum_{j=1}^{L_k}
 A^{\mathrm{VETTA}}_{k,j}&=\Aturn_k,\\
 A^{\mathrm{VETTA}}_{k,i}-A^{\mathrm{VETTA}}_{k,j}
 &=\alpha\left(\Atok_{k,i}-\Atok_{k,j}\right).
\end{align*}
This projection is an algebraic characterization of the scalar credit weights,
not an additional optimization step during training.
Different tokens have different score-function gradients, so zero scalar sum
does not imply a zero parameter-gradient contribution.

Under the additional special case
$\gamma_{\mathrm{tok}}=\lambda_{\mathrm{tok}}=1$, complete termination, no
token-wise KL reward, and reward only at an action's last valid token, all
prefixes in action $k$ share the same remaining sampled return $G_k$.  Then
\begin{equation}
 \Atok_{k,j}=G_k-\Vtok_{k,j},\qquad
 D_{k,j}=\frac{1}{L_k}\sum_{u=1}^{L_k}\Vtok_{k,u}-\Vtok_{k,j}.
\end{equation}
Within-response centering cancels the common sampled return. The resulting
token residual reflects differences among prefix value estimates, while
the turn advantage sets the response-level credit. This identity describes how credit varies across
positions in a sampled response; it does not identify the effect of replacing
an individual token.

\Needspace{10\baselineskip}
\subsection{Actor-loss implementation}
\label{app:actor-options}

Equation~\ref{eq:actor-update} gives the standard token-wise clipped PPO
surrogate. VETTA uses dual clipping for negative advantages. KL
regularization is enabled on ALFWorld, WebShop, and Sokoban, and disabled on
ScienceWorld. Dual clipping caps the per-token loss after PPO ratio
clipping at
$-c A^{\mathrm{VETTA}}_{k,j}$ when $A^{\mathrm{VETTA}}_{k,j}<0$, with $c=3$;
the loss for nonnegative advantages is unchanged. When enabled, KL
regularization is applied against a frozen reference policy initialized
from the actor's starting checkpoint. The ALFWorld and WebShop configurations
use the \texttt{low\_var\_kl} estimator and a KL coefficient of 0.01.
Both loss terms are averaged over valid generated-token positions. These
actor-loss settings are inherited from the token-wise PPO implementation
shared with the PPO and PPO* baselines; they are not specific to VETTA's
value estimation or advantage fusion.
The Turn-PPO baseline instead clips a whole-response probability ratio and
does not apply the negative-advantage dual-clipping rule.
\FloatBarrier

\end{document}